\documentclass[letterpaper,10pt,conference]{ieeeconf}

\def\hidedraftcomments{}

\IEEEoverridecommandlockouts
\usepackage{amsmath}
\usepackage{amssymb}
\usepackage{booktabs}
\usepackage{graphicx}
\usepackage{multirow}
\usepackage[table]{xcolor}
\usepackage{placeins}
\usepackage{tikz}
\usepackage{pgfplots}
\usepgfplotslibrary{groupplots}
\pgfplotsset{compat=1.18}
\usepackage{cite}
\usepackage{balance}
\usepackage[bookmarks=true,colorlinks=true,citecolor=blue,linkcolor=blue,urlcolor=blue]{hyperref}

\usepackage{subcaption}

\newcommand{\unmatchedbaseline}[1]{\cellcolor{black!10}#1}
\newcommand{\sr}[2]{\mbox{#1{\scriptsize$\,\pm\,$#2}}}
\newcommand{\bestsr}[2]{\mbox{\textbf{#1}{\scriptsize$\,\pm\,$\textbf{#2}}}}

\makeatletter
\let\NAT@parse\undefined
\makeatother
\usepackage[square, numbers, sort]{natbib}

\newif\ifshowcomments
\ifdefined\hidedraftcomments
    \showcommentsfalse
\else
    \showcommentstrue
\fi

\ifshowcomments
    \usepackage[textsize=scriptsize,textwidth=1.6cm]{todonotes}
\else
    \usepackage[disable,textsize=scriptsize]{todonotes}
\fi

\let\todonotesTodo\todo
\RenewDocumentCommand{\todo}{s O{} m}{%
    \IfBooleanTF{#1}
        {\todonotesTodo[inline,#2]{#3}}
        {\todonotesTodo[#2]{#3}}%
}

\NewDocumentCommand{\todohrc}{s O{} m}{%
    \IfBooleanTF{#1}
        {\todo*[fancyline,color=yellow!15,#2]{\textbf{HRC:} #3}}
        {\todo[fancyline,color=yellow!15,#2]{\textbf{HRC:} #3}}%
}
\NewDocumentCommand{\todojjt}{s O{} m}{%
    \IfBooleanTF{#1}
        {\todo*[fancyline,color=blue!15,#2]{\textbf{JJT:} #3}}
        {\todo[fancyline,color=blue!15,#2]{\textbf{JJT:} #3}}%
}
\NewDocumentCommand{\todokcs}{s O{} m}{%
    \IfBooleanTF{#1}
        {\todo*[fancyline,color=purple!15,#2]{\textbf{KCS:} #3}}
        {\todo[fancyline,color=purple!15,#2]{\textbf{KCS:} #3}}%
}
\NewDocumentCommand{\todomw}{s O{} m}{%
    \IfBooleanTF{#1}
        {\todo*[fancyline,color=green!15,#2]{\textbf{MW:} #3}}
        {\todo[fancyline,color=green!15,#2]{\textbf{MW:} #3}}%
}

\usepackage[most]{tcolorbox}

\newtcolorbox{takeaway}{
  colback=black!10,
  colframe=black!10,
  boxrule=0pt,
  arc=0pt,
  outer arc=0pt,
  left=3pt,
  right=3pt,
  top=2pt,
  bottom=2pt,
  boxsep=0pt,
  before skip=5pt,
  after skip=5pt,
}

\title{\LARGE \bf
CAP: Codebook-Aligned Prediction for Tokenized Robot Policies
}

\author{Haoran Chen$^{1,*}$, Jingtian Ji$^{1,*}$, Samuel Wheeler$^{2}$,\\
  Kaylene Caswell Stocking$^{1}$, and Matthew Walter$^{1}$%
  \thanks{$^{*}$Equal contribution.}%
  \thanks{$^{1}$Toyota Technological Institute at Chicago.}%
  \thanks{$^{2}$Argonne National Laboratory.}%
}

\hypersetup{
  pdftitle={CAP: Codebook-Aligned Prediction for Tokenized Robot Policies},
  pdfauthor={Haoran Chen, Jingtian Ji, Samuel Wheeler, Kaylene Caswell Stocking, Matthew Walter}
}

\begin{document}

\makeatletter
\IEEEaftertitletext{%
  \begin{minipage}{\textwidth}
    \centering
    \includegraphics[width=0.9\linewidth]{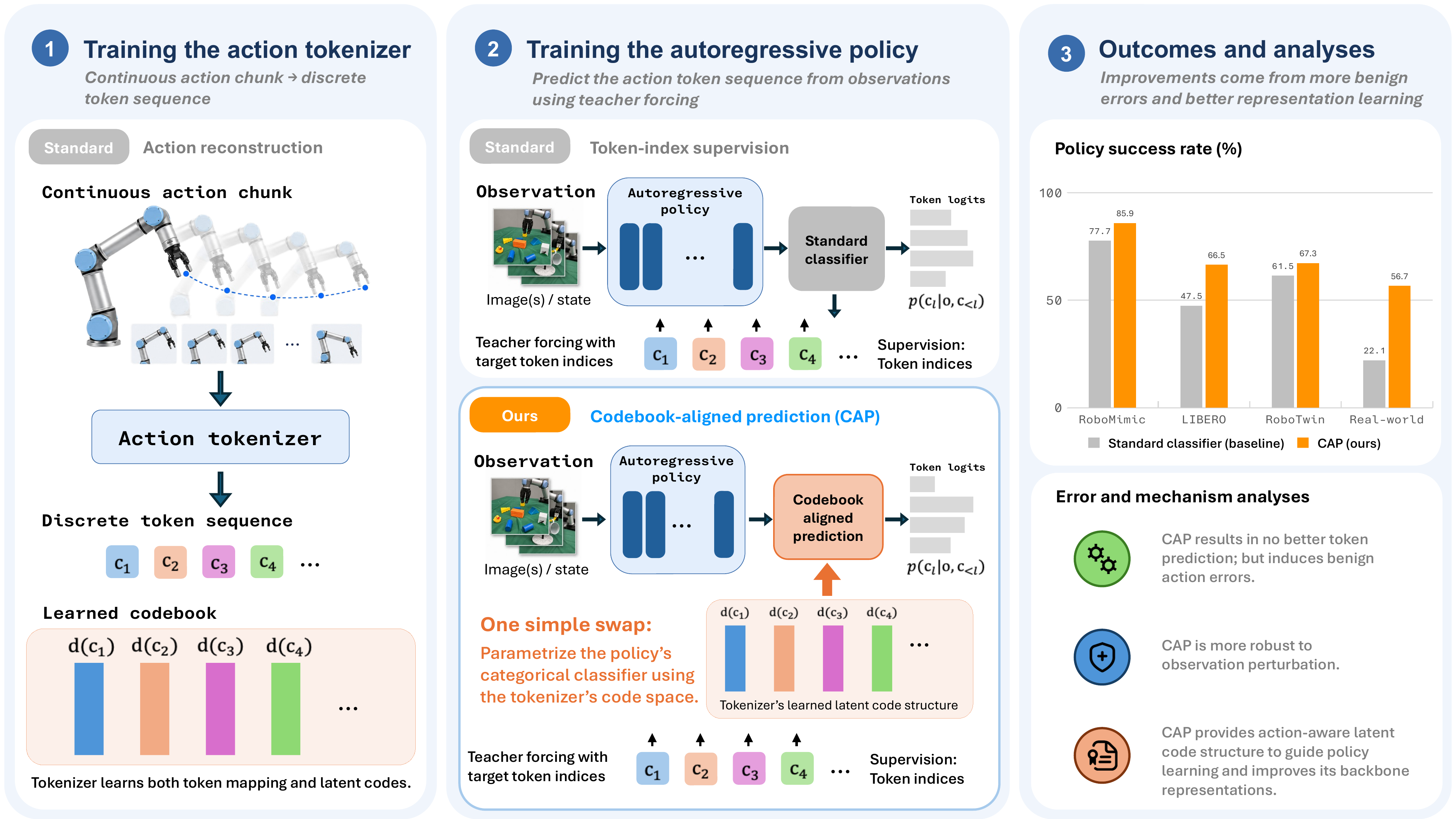}
    \def\@captype{figure}%
    \caption{\textbf{Codebook-Aligned Prediction (CAP).} An action tokenizer converts continuous action chunks into discrete token sequences while learning structured latent code representations. Standard autoregressive policies discard this structure and predict token indices with a learnable linear head. CAP instead reuses the tokenizer’s codebook to structure the policy prediction space. This simple replacement yields more benign action-space errors and better policy representations, improving downstream success across simulation and real-world tasks.}
    \label{fig:teaser}
  \end{minipage}%
}
\makeatother

\maketitle
\thispagestyle{empty}
\pagestyle{empty}

\begin{abstract}
Action tokenization converts continuous robot actions into discrete symbols that can be modeled autoregressively. 
However, existing tokenizer-based policies typically ignore the tokenizer's learned latent code structure:
after tokenization, the policy treats tokens as unrelated class indices and learns a new classifier from scratch.
We show that this discarded structure is valuable. We introduce \emph{Codebook-Aligned Prediction} (CAP), a method that directly reuses the tokenizer's code vectors as policy class prototypes while leaving the tokenizer and policy backbone otherwise unchanged. 
Across four quantizer families, three simulation benchmarks, and two real-robot tasks, CAP consistently improves task success over standard token classification heads while holding the tokenizer---and therefore its reconstruction quality---fixed. 
Our analysis further shows that these gains are not explained by 
higher token accuracy or changes in the policy head alone.
Instead, reusing the tokenizer codebook provides the policy with valuable information about the tokenizer's learned latent structure across tokens, 
making token prediction errors more benign in action space 
and improving the representations learned by the policy backbone. 
These results suggest that action tokenizers learn useful action-aware latent
structure beyond discrete targets that should be preserved when training downstream policies. 
\end{abstract}

\section{Introduction}
\label{sec:introduction}

Robot policies learned from demonstrations now range from task-specific visuomotor controllers~\cite{Zhao2023ACT,Chi2023DiffusionPolicy} to generalist vision--language--action (VLA) models~\cite{Kim2024OpenVLA,intelligence2025pi05}. 
A critical design choice for
robot policies is how to represent the action space. 
Common approaches include direct regression~\cite{Haldar2024BAKU}, continuous generative models~\cite{Zhao2023ACT, Chi2023DiffusionPolicy, Chisari2025PointFlowMatch}, and discrete autoregressive models~\cite{Shafiullah2022BehaviorTransformer,yunis24a}.
Each formulation presents distinct challenges for robot policies, including differences in training objectives and inference procedures.

In this work, we focus on the last formulation: action tokenizers that discretize continuous action chunks, followed by autoregressive policies that predict the resulting action tokens.
Casting action prediction as categorical next-token prediction enables the policy to represent multimodal action distributions, provide explicit token likelihoods, and avoid iterative continuous-sample refinement at inference time.
For VLA models, it allows the policy to generate actions using the same autoregressive framework employed for language~\cite{Zitkovich2023RT2,Pertsch2025FAST,Wang2025VQVLA}. 
The tokenizer does more than simply convert continuous actions into discrete tokens for the policy: it determines reconstruction fidelity, sequence length, and the target distribution that the policy must learn.


Existing action tokenizers fall into two broad categories.
Predefined tokenizers map actions to tokens using fixed coordinates or trajectory transforms, including per-dimension bins~\cite{Brohan2023RT1}, frequency coefficients~\cite{Pertsch2025FAST}, and B-spline bases~\cite{Zhou2025BEAST}.
In contrast, learned action tokenizers adapt the discrete representation to the action distribution. Among these approaches, VQ-BeT~\cite{Lee2024VQBeT} and QueST~\cite{Mete2024QueST} learn latent vocabularies; VQ-VLA~\cite{Wang2025VQVLA} and FASTer~\cite{Liu2025FASTer} emphasize reconstruction and compression for autoregressive VLA models; OAT~\cite{Liu2026OAT} imposes coarse-to-fine structure across tokens; and other methods further shape the vocabulary through optimization-oriented or multimodal objectives~\cite{Dong2026ActionCodec,Kang2026XTokenizer}.

Once trained, action tokenizers are typically used only to provide discrete token labels for downstream policy learning, while their associated code representations are discarded. This creates a mismatch between the structure learned by the tokenizer and how its tokens are modeled by the policy. In multi-level tokenizers, for example, the same token index may correspond to different code representations across levels, yet standard policies commonly use a shared classifier across the token sequence. More generally, one-hot supervision penalizes all incorrect tokens equally, regardless of their similarity to the target action. Consequently, the policy prediction objective ignores information already encoded in the tokenizer's latent code space. This raises a natural question: can these learned representations be used to better inform downstream policy learning?

Our work answers this question affirmatively. We propose \emph{Codebook-Aligned Prediction} (CAP) for policy training, which reuses tokenizer code vectors as class prototypes in the policy classifier and scores output representations by their distance to these prototypes. 
CAP therefore inherently aligns the policy's prediction heads with the tokenizer's level-wise structure and retains information already learned by the tokenizer without changing the tokenizer, policy backbone, or autoregressive training objective.

Our contributions are summarized below:
\begin{itemize}
    \setlength{\topsep}{2pt}
    \setlength{\itemsep}{0.5pt}
    \setlength{\parsep}{0pt}
    \setlength{\parskip}{0pt}
    %
    \item We propose \textbf{\emph{Codebook-Aligned Prediction} (CAP)}, a simple prediction mechanism that reuses rather than discards tokenizer code representations and that can be used with both learned-codebook and lookup-free quantizers.
    \item We find that \textbf{CAP consistently improves task success} across four quantizer families, three simulation benchmarks, and two real-world tasks.

    \item We perform \textbf{thorough analysis and ablations} to investigate the causes of CAP's effectiveness, attributing these gains to more benign action errors and improved policy representations.
\end{itemize}

\section{Related Work}\label{sec:related-work}


\subsection{Quantization and Robot Action Tokenizers}

Quantization has long been used for discrete representation learning, from scalar and vector quantization~\cite{Gray1984VectorQuantization} to neural quantizers such as VQ-VAE~\cite{Oord2017VQVAE}, residual quantization~\cite{Lee2022RQ}, and structured designs like FSQ~\cite{Mentzer2024FSQ} 
and BSQ~\cite{Zhao2024BSQ}. Its applications span vision~\cite{Esser2021VQGAN}, audio~\cite{Zeghidour2022SoundStream}, and speech~\cite{Hsu2021HuBERT}.


%

In robotics, action tokenizers use either predefined or learned representations.
Predefined quantization schemes discretize action dimensions or transformed trajectories, as in RT-1~\cite{Brohan2023RT1}, FAST~\cite{Pertsch2025FAST}, and related approaches~\cite{Zhou2025BEAST}. 
Frequency- and spline-based methods~\cite{Pertsch2025FAST,Zhou2025BEAST} alleviate the redundancy of simple binning methods with compact trajectory representations, but their underlying bases remain fixed by design rather than adapting to correlations or recurring patterns in the action data.
Learned tokenizers instead adapt their representations to the action distribution: VQ-BeT and QueST learn latent action vocabularies~\cite{Lee2024VQBeT,Mete2024QueST}, VQ-VLA~\cite{Wang2025VQVLA} and FASTer~\cite{Liu2025FASTer} integrate learned tokens into autoregressive VLA models, and recent methods further structure the vocabulary through architectural and auxiliary objectives~\cite{Dong2026ActionCodec,Liu2026OAT,Kang2026XTokenizer}.
These works improve how actions are tokenized, but the downstream policy only receives token index labels and has to learn a new categorical classifier from scratch, ignoring the prior knowledge in the tokenizer's learned latent structure.
In this work, we study how this structure can instead be leveraged to improve the policy.

\subsection{Imitation Learning}
Imitation learning trains policies with expert demonstrations, typically through behavior cloning, and is widely used for robot manipulation~\cite{Ross2011DAgger}. Action chunking predicts a sequence of future actions rather than a single action, allowing policies to capture short-horizon temporal structure and reduce the frequency of policy queries. Among methods that use action chunking, 
different policy parameterizations model the action distribution in different ways. ACT~\cite{Zhao2023ACT} directly predicts continuous action chunks with a conditional VAE, Diffusion Policy~\cite{Chi2023DiffusionPolicy} learns a denoising process, and flow-matching policies~\cite{Chisari2025PointFlowMatch} regress a conditional vector field.
For tokenized policies, the action chunks are instead represented as discrete token sequences and predicted autoregressively~\cite{Shafiullah2022BehaviorTransformer,Lee2024VQBeT,Mete2024QueST}.
We focus on this formulation and study how token representation alignment affects policy behavior.

\section{Method} \label{sec:method}

We first formulate the standard two-stage training pipeline where a tokenizer first learns a mapping between continuous actions and discrete tokens and an autoregressive policy then learns to predict the appropriate action tokens. 
We then identify two structural mismatches at their interface: (1) the same token index may correspond to different code representations across levels in a tokenizer, yet standard policies predict all levels with a shared categorical head, and (2) policy training uses only the discrete token indices and ignores the latent structure learned by the tokenizer.
To address these issues, we propose \emph{Codebook-Aligned Prediction} (CAP) to structure the policy's prediction space by reusing tokenizer representations as class prototypes for policy training.

\subsection{Preliminaries}
\label{sec:prelim}
In this work, an action tokenizer first maps a continuous action chunk to a sequence of discrete tokens and learns to reconstruct the original actions.
The trained tokenizer is then frozen, and its token sequences serve as categorical targets for an autoregressive policy conditioned on observations.

\textbf{Action Tokenizer:}
Let $\mathbf a=(a_1,\ldots,a_H)\in\mathbb R^{H\times d_a}$ denote an $H$-step action chunk, where $a_t\in\mathbb R^{d_a}$.
A tokenizer consists of an encoder $E$, a quantizer $Q=(q,d)$, and an action decoder $D$. The encoder first maps the action chunk to a continuous embedding, $z_e=E(\mathbf a)$. The indexing map $q$ then converts $z_e$ into a sequence of discrete action tokens, $\mathbf c=q(z_e):=(c_1,\ldots,c_L)\in[K]^L$, where $c_l\in[K]=\{1,\ldots,K\}$ is the token index at level $l$, $L$ is the number of quantization levels, and $K$ is the number of possible indices per level. The index-to-latent map $d$ converts these indices into a continuous quantized embedding, $z_q=d(\mathbf c)$, using the corresponding level-wise codes. Finally, the decoder reconstructs the action chunk from the quantized embedding, i.e., $\mathbf a'=D(z_q)$.

We therefore define tokenization and detokenization as $\mathcal T(\mathbf a):=q(E(\mathbf a))$ and $\mathcal D(\mathbf c):=D(d(\mathbf c))$, respectively. A tokenizer is typically trained with a reconstruction objective:
\begin{equation}
    \mathcal L_{\mathrm{tok}}
    =\mathbb E_{\mathbf a\sim p_{\mathrm{data}}(\mathbf a)}\!\left[
      \ell_{\mathrm{rec}}\!\left(\mathbf a,\mathcal D(\mathcal T(\mathbf a))\right)
    \right].
    \label{eq:tokenizer-objective}
\end{equation}
Here, $p_{\mathrm{data}}(\mathbf a)$ denotes the action-chunk distribution in the demonstrations. We use the mean squared error (MSE) to define the reconstruction loss:
\begin{equation}
    \ell_{\mathrm{rec}}(\mathbf a,\mathbf a^{\prime})
    =\frac{1}{Hd_a}\sum_{t=1}^{H}\lVert a_t- a^{\prime}_t\rVert_2^2,
    \label{eq:reconstruction-error}
\end{equation}
For VQ-based tokenizers, we additionally use the standard codebook and commitment losses.

\textbf{Autoregressive Token Policy:}
Let $p_{\mathrm{data}}(o,\mathbf a)$ denote the demonstration distribution, where $o$ is the observation context paired with action chunk $\mathbf a$.
Once the tokenizer has been trained, we freeze it and map each action chunk to a target token sequence \mbox{$\mathbf c=\mathcal T(\mathbf a)\in[K]^L$.}

A standard policy with parameters $\psi$ uses a learned classifier shared across levels to predict categorical logits and models this sequence autoregressively~\cite{Shafiullah2022BehaviorTransformer, Lee2024VQBeT,Mete2024QueST,Wang2025VQVLA}:
\begin{equation}
    \begin{aligned}
        p_\psi(\mathbf c\mid o)
          &=\prod_{l=1}^{L}p_\psi(c_l\mid c_{<l},o),\\
        \mathcal L_{\mathrm{AR}}
          &=-\mathbb E_{(o,\mathbf a)\sim p_{\mathrm{data}}}\sum_{l=1}^{L}
            \log p_\psi(c_l\mid c_{<l},o).
    \end{aligned}
    \label{eq:ar-objective}
\end{equation}
Here $c_{<l}:=(c_1,\ldots,c_{l-1})$ is the ground-truth sequence prefix. The policy is trained by minimizing the token-level negative log-likelihood through teacher forcing.
During a rollout, the policy generates $\widehat{\mathbf c}=(\widehat c_1,\ldots,\widehat c_L)$ autoregressively, and the frozen detokenizer reconstructs the predicted action chunk as $\widehat{\mathbf a}:=\mathcal D(\widehat{\mathbf c})$.

\subsection{Multi-Level Quantization and Policy Mismatch}
\label{sec:multi-level}

\textbf{Multi-level Quantizers.} The forms of the indexing and index-to-latent maps, $q$ and $d$, depend on the quantizer family. We consider four representative multi-level quantizers spanning both explicit learned codebooks and lookup-free quantization, as well as parallel and sequential residual constructions.
Vector quantization (VQ) and finite scalar quantization (FSQ) quantize $L$ latent vectors in \emph{parallel}; residual VQ (RVQ) and normalized RVQ (NRVQ) quantize a single latent vector sequentially through $L$ \emph{residual} stages.

For VQ, the encoder produces $L$ latent vectors $z_e=(z_{e,1},\ldots,z_{e,L})\in\mathbb R^{L \times d_c}$. At level $l$, let
$\mathcal E_l=\{v_{l,k}\}_{k=1}^{K}$ denote the codebook, with $v_{l,k}\in\mathbb R^{d_c}$. Each latent vector is independently quantized as
\begin{equation}
    c_l=\arg\min_{k\in[K]}
    \|z_{e,l}-v_{l,k}\|_2^2,
    \qquad
    z_{q,l}=v_{l,c_l}.
    \label{eq:lookup}
\end{equation}
For RVQ and NRVQ, the encoder generates a single latent $z_e\in\mathbb R^{d_c}$. 
RVQ applies the lookup sequentially to the residual with $r_0=z_e$,
\begin{equation}
    c_l=\arg\min_{k\in[K]}
    \|r_{l-1}-v_{l,k}\|_2^2,
    \qquad
    r_l=r_{l-1}-v_{l,c_l},
    \label{eq:rvq}
\end{equation}
and the quantized latent is
$z_q=\sum_{l=1}^{L}v_{l,c_l}$~\cite{Lee2022RQ,Zeghidour2022SoundStream}.
NRVQ follows the same residual structure but performs assignment using normalized code directions $\bar v_{l,k}=v_{l,k}/(\|v_{l,k}\|_2+\epsilon)$ and scales the selected direction at each level by a learned scalar $s_l$ before reconstruction~\cite{Yu2022ImprovedVQGAN}.

FSQ uses a fixed Cartesian grid instead of a learned lookup table for the $L$ encoded latent vectors~\cite{Mentzer2024FSQ}. Let $\phi$ map grid points to token indices; its quantized latent at level $l$ is
\begin{equation}
    c_l=\phi\left(
    \Pi_l^{\mathrm{in}} (z_{e,l})
    \right),
    \qquad
    z_{q,l}
    =\Pi_l^{\mathrm{out}}\!\left(\phi^{-1}(c_l)\right),
    \label{eq:fsq}
\end{equation}
where $\Pi_l^{\mathrm{in}}$ is a linear layer with $\tanh$-based bounding and $\Pi_l^{\mathrm{out}}$ maps the selected grid point to the decoder embedding. Its finite Cartesian-product code space is closely related to product quantization~\cite{Jegou2011ProductQuantization}.

\textbf{Shared and Level-Specific Quantization.}
Let $\Theta_l$ denote the quantization parameters used at level $l$. A \emph{shared} quantizer ties these parameters across levels, $\Theta_l=\Theta$, whereas a \emph{level-specific} quantizer uses independently parameterized $\Theta_l$. For VQ and RVQ, $\Theta_l$ corresponds to the codebook $\mathcal E_l$; for NRVQ, to the normalized direction table; and for FSQ, to the projection pair $(\Pi_l^{\mathrm{in}},\Pi_l^{\mathrm{out}})$. Both parameterizations nevertheless produce token sequences in the same discrete space $\mathbf c\in[K]^L$.

\textbf{Tokenizer--Policy Mismatch.}
A standard autoregressive policy (Eqn.~\ref{eq:ar-objective}) typically uses a single head (\textsc{SH}), with one categorical classifier shared across quantization levels. This is potentially problematic for level-specific quantizers, where the same token index can correspond to different code representations across levels: a shared head must use one set of classifier parameters for token classes whose tokenizer-side meanings are level-dependent.

A simple remedy is to match the policy-head structure to the tokenizer. We therefore also consider a multi-head policy (\textsc{MH}), with a separate categorical classifier per level. We call \textsc{SH} structurally \emph{matched} to shared quantization and \textsc{MH} \emph{matched} to level-specific quantization; we refer to the opposite pairings as \emph{mismatched}. Such matching provides a natural inductive bias, preserves the tokenizer's parameter-sharing structure in the downstream policy, and avoids unnecessary coupling between independently parameterized levels.

\subsection{Codebook-Aligned Prediction}
\label{sec:cap}
Structural matching addresses how prediction heads are shared across levels, but it does not use the latent structure learned by the tokenizer. Even with matched heads, categorical supervision treats token indices as discrete class labels and ignores action-relevant relationships encoded in their latent representations which have been learned while training the tokenizer. We therefore propose \emph{Codebook-Aligned Prediction} (CAP) to target this remaining gap by carrying that tokenizer structure into policy prediction.

\textbf{Class prototype construction:}
At level $l$, CAP associates each token index $k$ with a tokenizer-derived code representation $V_l[k]\in\mathbb R^{d_c}$, forming a class prototype matrix $V_l\in\mathbb R^{K\times d_c}$. For quantizers with explicit learned codebooks, we reuse their code representations directly. For lookup-free quantizers, we provide a systematic way to construct their corresponding implicit codebooks.

\paragraph{Learned Codebooks}

For VQ and RVQ, CAP directly reuses the learned code vectors, $V_l[k]=v_{l,k}$ (Eqn.~\ref{eq:lookup}). For NRVQ, we use the normalized directions, $V_l[k]=\bar v_{l,k}$.

\paragraph{Implicit Codebooks}


Lookup-free quantizers replace learned codebooks with structured finite code sets.
These code sets define an implicit codebook. For FSQ, we enumerate the grid points and map them through the decoder-side projection to construct CAP's prototypes:
\begin{equation}
    V_l[k]
    =\Pi_l^{\mathrm{out}}\!\left(\phi^{-1}(k)\right),
    \qquad k\in[K].
    \label{eq:fsq-codebook}
\end{equation}
Thus, the resulting prototypes are decoder-input embeddings rather than grid coordinates.
Note that the same construction can be extended to LFQ and BSQ using their respective binary code sets and decoder-side embedding maps.

\textbf{Prototype-based token prediction:}
CAP, by construction, follows the structurally matched head configuration introduced above: shared quantization uses a single head (\textsc{SH}) across levels, whereas level-specific quantization uses one head per level (\textsc{MH}). This ensures that each policy head is paired with the code representations that define the corresponding tokenizer level.

At level $l$, let $h_l\in\mathbb R^{d_{\mathrm{model}}}$ denote the policy embedding for token prediction at position $l$.
We learn an affine down-projection $P_l:\mathbb R^{d_{\mathrm{model}}}\to\mathbb R^{d_c}$ and a positive logit scale $\tau_l$.
$P_l$, $V_l$, and $\tau_l$ are tied across levels for shared quantization and learned independently for level-specific quantization.
With CAP, the policy output embedding is translated into a distribution over action tokens via:
\begin{equation}
    \begin{aligned}
        u_l=P_l(h_l),
        \qquad
        \eta_{l,k}=-\tau_l\bigl\lVert u_l-V_l[k]\bigr\rVert_2^2 \\
        p(c_l=k\mid c_{<l},o)=\operatorname{softmax}(\eta_{l,\cdot})_k,
    \end{aligned}
    \label{eq:distance-logits}
\end{equation}
where $\eta_{l,k}$ denotes the logit for predicting token $k$. Tokens whose codebook latent representations $V_l[k]$ are closer to the projected policy output $u_l$ are assigned higher probability.

\textbf{Frozen and fine-tuned CAP:}
We consider two variants. \emph{Frozen CAP} (\textsc{CAP-frz}) keeps the prototypes fixed at their tokenizer-derived values; \emph{Fine-tuned CAP} (\textsc{CAP-ft}) updates them through the policy cross-entropy objective.
The tokenizer itself remains fixed during policy training. 

\begin{figure}[!t]
  \centering
  \captionsetup[sub]{font=scriptsize}
  \begin{subfigure}[t]{0.33\linewidth}
    \includegraphics[width=\linewidth]{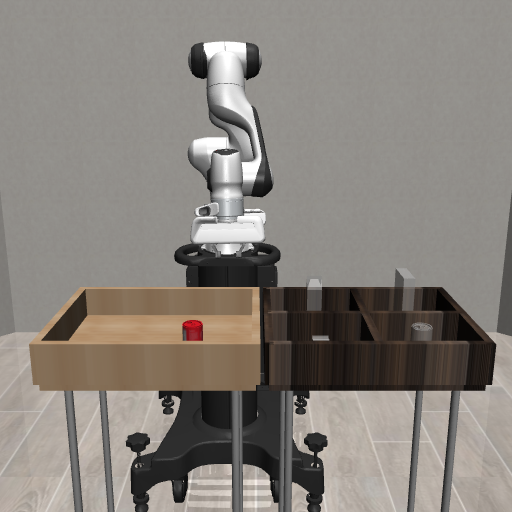}
    \caption{RoboMimic: Can}
  \end{subfigure}\hfil
  \begin{subfigure}[t]{0.33\linewidth}
    \includegraphics[width=\linewidth]{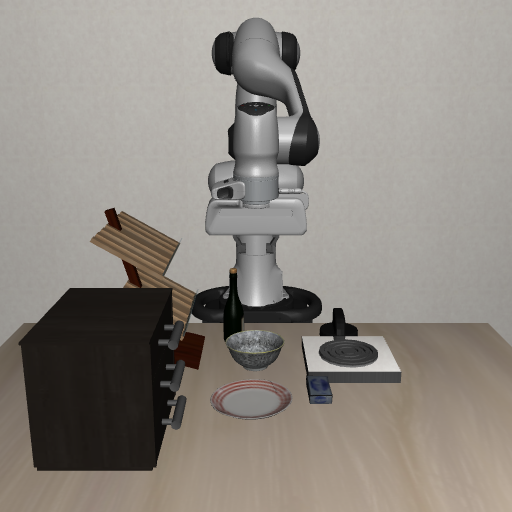}
    \caption{LIBERO: Goal}
  \end{subfigure}\hfil
  \begin{subfigure}[t]{0.33\linewidth}
    \includegraphics[width=\linewidth]{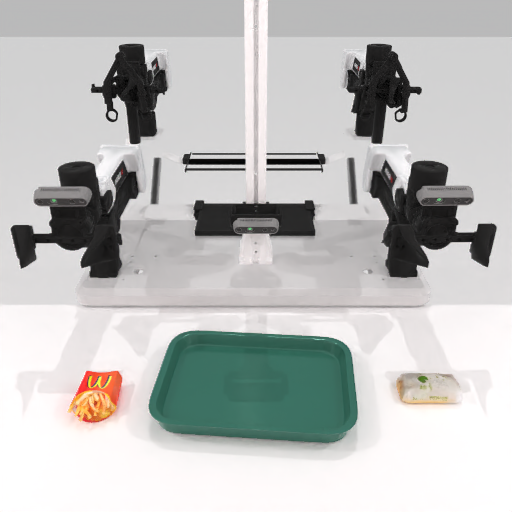}
    \caption{RoboTwin: Burger\&Fries}
  \end{subfigure}
  \caption{Example tasks from the three simulation benchmarks.}
  \label{fig:simulation-benchmarks}
  \vspace{-5px}
\end{figure}

\section{Results} \label{sec:experiments}

\subsection{Experimental Setup} \label{sec:setting}

\begin{figure}[!t]
  \centering
  \begin{subfigure}{0.45\linewidth}
    \includegraphics[width=\linewidth]{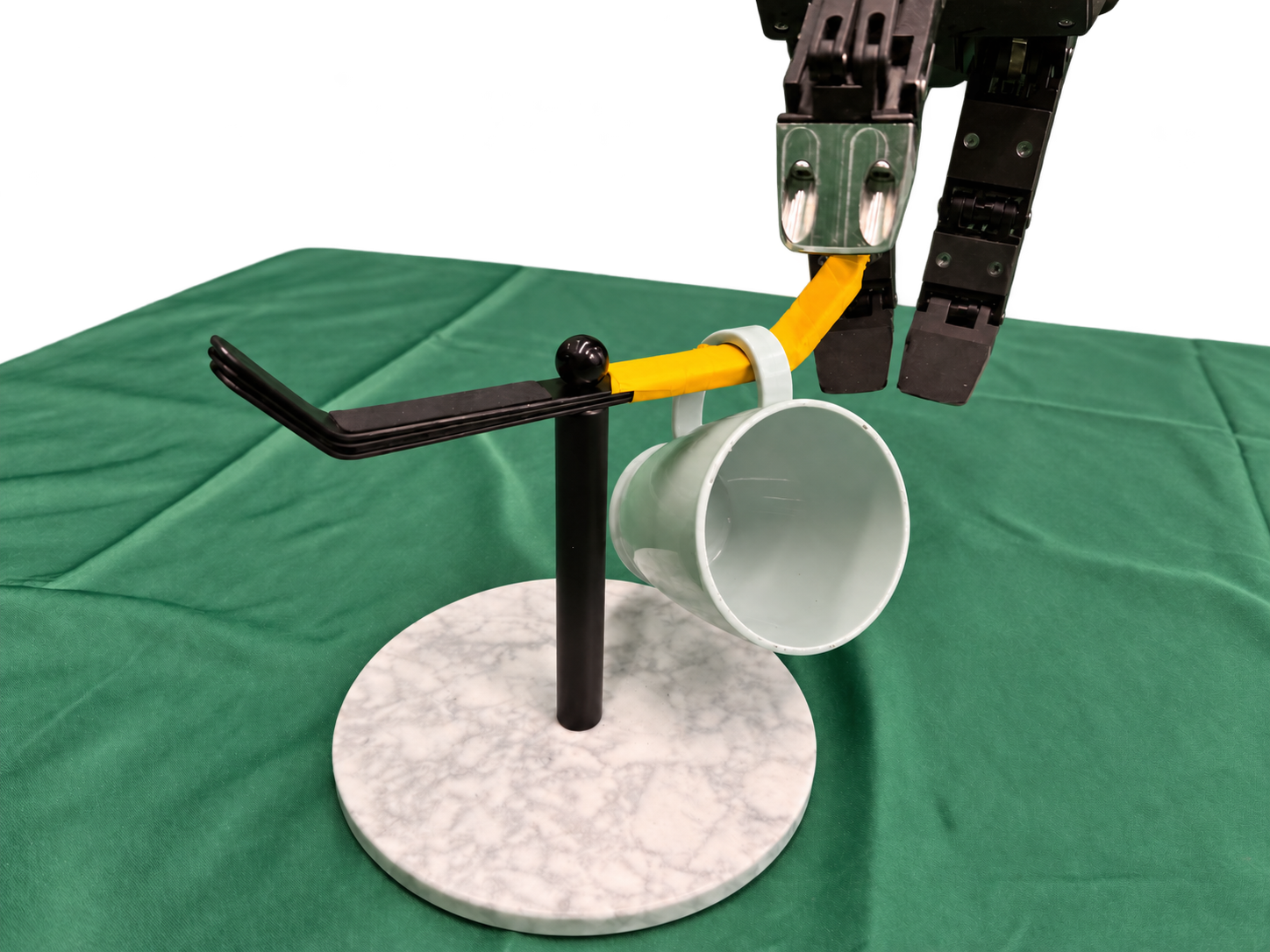}
    \caption{Cup Hang}\label{fig:cup-hang}
  \end{subfigure}\hfil
  \begin{subfigure}{0.45\linewidth}
    \includegraphics[width=\linewidth]{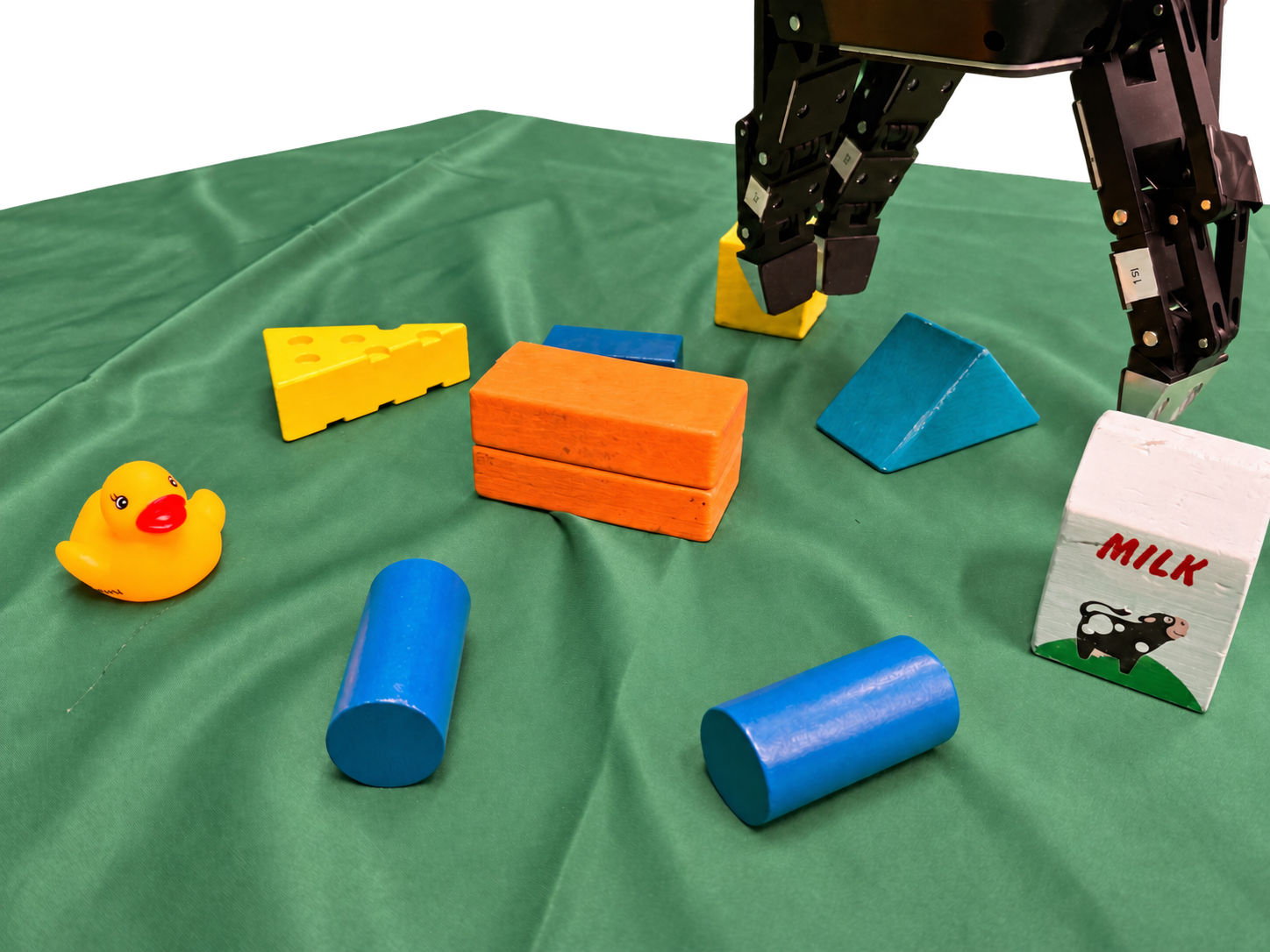}
    \caption{Cube Stack}\label{fig:cube-stack}
  \end{subfigure}
  \caption{Real-world evaluation tasks.}\label{fig:real-world-tasks}
  \vspace{-0.25em}
\end{figure}

\paragraph{Benchmarks and Evaluation}
We evaluate CAP on three simulation benchmarks and two real-world tasks spanning single-arm and bimanual manipulation.
For the simulation benchmarks (Fig.~\ref{fig:simulation-benchmarks}), we use Lift, Square, and Can from RoboMimic~\cite{Mandlekar2022RoboMimic}; the Spatial, Object, Goal, and Long suites from LIBERO~\cite{Liu2023LIBERO}, with 10 tasks per suite (40 tasks in total); and Grab Roller, Place Burger and Fries, and Place Cans in a Plastic Box from RoboTwin~\cite{Chen2025RoboTwin2}. The real-world tasks (Fig.~\ref{fig:real-world-tasks}) 
involve hanging a coffee cup on a rack (Cup Hang) and stacking a cube (Cube Stack). 
We train a separate tokenizer and policy for each simulation benchmark using all tasks within that benchmark. For the real-world setup, we likewise train one tokenizer and one policy jointly on the combined two-task dataset.
We evaluate each policy on $300$ episodes per RoboMimic task, $50$ per LIBERO task using the benchmark's provided evaluation seeds, $100$ per RoboTwin task, and $30$ per real-world task. 
We report the policy's rollout success rate along with the bootstrap estimate of the standard error, using the validation-selected checkpoint for simulation environments and the final checkpoint for the real-world experiments.

\paragraph{Quantizer Configurations}
We evaluate VQ, RVQ, NRVQ, and FSQ with both shared and level-specific parameterizations as described in Sec.~\ref{sec:multi-level}, denoted by the suffixes \textsc{-S} and \textsc{-L}
, respectively.
For all experiments, we set $H = 8$, $L = 8$, and $K = 128$.

\paragraph{Baselines and Our Methods}
For each tokenizer configuration, we train both single-head (\textsc{SH}) and multi-head (\textsc{MH}) 
baselines along with \textsc{CAP-frz} and \textsc{CAP-ft}.
All policy variants use the same transformer backbone, frozen tokenizer, training data, and optimization protocol; only the prediction heads differ.


\subsection{Simulation Results} \label{sec:main-results}

\begin{table*}[t]
  \centering
  \caption{Simulation results: raw-action reconstruction MSE
  (RoboMimic/LIBERO: $\times10^{-3}$; RoboTwin: $\times10^{-6}$)
  and success rate (\%, $\pm$ bootstrap standard error).
  Gray: mismatched baseline; bold: best policy for each benchmark--tokenizer pair.}
  \label{tab:main}
  \scriptsize
  \renewcommand{\arraystretch}{0.92}
  \setlength{\tabcolsep}{1pt}
  \begin{tabular}{l c cccc @{\hspace{3pt}} c cccc @{\hspace{3pt}} c cccc}
    \toprule
    & \multicolumn{5}{c}{RoboMimic} & \multicolumn{5}{c}{LIBERO} & \multicolumn{5}{c}{RoboTwin} \\
    \cmidrule(lr){2-6}\cmidrule(lr){7-11}\cmidrule(lr){12-16}
    Quantizer
      & Recon.$\downarrow$ & \textsc{SH} & \textsc{MH} & \textsc{CAP-frz} & \textsc{CAP-ft}
      & Recon.$\downarrow$ & \textsc{SH} & \textsc{MH} & \textsc{CAP-frz} & \textsc{CAP-ft}
      & Recon.$\downarrow$ & \textsc{SH} & \textsc{MH} & \textsc{CAP-frz} & \textsc{CAP-ft} \\
    \midrule
    VQ-S
      & 2.02 & \sr{80.2}{1.3} & \unmatchedbaseline{\sr{77.2}{1.4}} & \bestsr{89.8}{1.0} & \sr{89.7}{1.0}
      & 0.600 & \sr{45.4}{1.1} & \unmatchedbaseline{\sr{17.4}{0.9}} & \bestsr{61.5}{1.1} & \sr{61.2}{1.1}
      & 8.90 & \sr{75.7}{2.5} & \unmatchedbaseline{\sr{53.7}{2.9}} & \bestsr{88.7}{1.8} & \sr{86.0}{2.0} \\
    VQ-L
      & 2.17 & \unmatchedbaseline{\sr{71.9}{1.5}} & \sr{78.8}{1.4} & \sr{91.7}{0.9} & \bestsr{92.0}{0.9}
      & 0.954 & \unmatchedbaseline{\sr{49.4}{1.1}} & \sr{51.4}{1.1} & \bestsr{74.2}{1.0} & \sr{71.0}{1.0}
      & 4.47 & \unmatchedbaseline{\sr{39.3}{2.8}} & \sr{53.0}{2.9} & \bestsr{93.0}{1.5} & \sr{92.0}{1.6} \\
    FSQ-S
      & 2.74 & \sr{74.8}{1.4} & \unmatchedbaseline{\sr{75.3}{1.4}} & \sr{85.3}{1.2} & \bestsr{85.8}{1.2}
      & 1.35 & \sr{32.2}{1.0} & \unmatchedbaseline{\sr{23.5}{0.9}} & \bestsr{62.3}{1.1} & \sr{50.5}{1.1}
      & 8.10 & \sr{69.7}{2.6} & \unmatchedbaseline{\sr{66.7}{2.7}} & \sr{75.7}{2.5} & \bestsr{84.0}{2.1} \\
    FSQ-L
      & 2.46 & \unmatchedbaseline{\sr{74.0}{1.5}} & \sr{79.4}{1.4} & \bestsr{88.4}{1.1} & \sr{87.9}{1.1}
      & 1.22 & \unmatchedbaseline{\sr{16.7}{0.8}} & \sr{25.0}{1.0} & \bestsr{71.1}{1.0} & \sr{61.6}{1.1}
      & 6.39 & \unmatchedbaseline{\sr{67.0}{2.7}} & \sr{65.3}{2.8} & \sr{84.7}{2.1} & \bestsr{85.3}{2.0} \\
    RVQ-S
      & 3.37 & \sr{78.4}{1.4} & \unmatchedbaseline{\sr{87.9}{1.1}} & \sr{86.7}{1.1} & \bestsr{88.2}{1.1}
      & 1.58 & \sr{64.5}{1.1} & \unmatchedbaseline{\sr{47.8}{1.1}} & \bestsr{66.4}{1.0} & \sr{49.3}{1.1}
      & 7.31 & \bestsr{72.3}{2.6} & \unmatchedbaseline{\bestsr{72.3}{2.6}} & \sr{59.3}{2.8} & \sr{56.3}{2.9} \\
    RVQ-L
      & 2.43 & \unmatchedbaseline{\sr{55.6}{1.6}} & \sr{74.4}{1.5} & \sr{80.6}{1.3} & \bestsr{80.7}{1.3}
      & 1.01 & \unmatchedbaseline{\sr{51.9}{1.1}} & \sr{55.2}{1.1} & \bestsr{64.6}{1.1} & \sr{64.4}{1.1}
      & 4.53 & \unmatchedbaseline{\sr{26.3}{2.5}} & \sr{45.7}{2.9} & \sr{31.3}{2.7} & \bestsr{63.7}{2.8} \\
    NRVQ-S
      & 4.31 & \sr{80.7}{1.3} & \unmatchedbaseline{\sr{83.3}{1.2}} & \sr{80.8}{1.3} & \bestsr{87.0}{1.1}
      & 1.74 & \sr{38.7}{1.1} & \unmatchedbaseline{\sr{50.4}{1.1}} & \sr{62.9}{1.1} & \bestsr{64.4}{1.1}
      & 13.8 & \sr{63.3}{2.8} & \unmatchedbaseline{\sr{70.7}{2.6}} & \sr{48.0}{2.9} & \bestsr{76.0}{2.5} \\
    NRVQ-L
      & 2.66 & \unmatchedbaseline{\sr{48.3}{1.7}} & \sr{74.9}{1.4} & \bestsr{84.0}{1.2} & \sr{81.9}{1.3}
      & 1.37 & \unmatchedbaseline{\sr{49.1}{1.1}} & \sr{67.8}{1.0} & \bestsr{69.2}{1.0} & \sr{68.3}{1.0}
      & 8.34 & \unmatchedbaseline{\sr{34.0}{2.7}} & \sr{47.0}{2.9} & \sr{57.7}{2.8} & \bestsr{58.3}{2.8} \\
    \bottomrule
  \end{tabular}
  \vspace{-5px}
\end{table*}

Table~\ref{tab:main} 
reports validation raw-action reconstruction MSE and policy success rate on the simulation benchmarks,
where gray cells indicate the \emph{mismatched} baseline.
The use of multiple policy heads is most consequential for level-specific tokenizers, where a shared policy head must serve independently learned codebooks. 
Accordingly, \textsc{MH} outperforms \textsc{SH} in eleven of the twelve level-specific comparisons; 
the only exception is RoboTwin FSQ-L, where \textsc{SH} is higher by $1.7$ percentage points. 
For shared tokenizers, the comparison depends on the quantizer type. \textsc{MH} outperforms \textsc{SH} in four of six RVQ-S and NRVQ-S comparisons. In contrast, \textsc{SH} outperforms \textsc{MH} in five of six VQ-S and FSQ-S comparisons.
We hypothesize that the benefit of multiple heads for shared \emph{residual} tokenizers arises from the stage-specific roles of tokens: even with a shared codebook, different levels encode progressively finer residual information and may therefore benefit from level-specific classifiers (\textsc{MH}). 

Across all three benchmarks, a CAP variant outperforms both baselines for all but a single (one out of 24) benchmark-tokenizer combination. 
Relative to the matched vanilla baseline (\textsc{SH} for \textasteriskcentered-S and \textsc{MH} for  \textasteriskcentered-L), \textsc{CAP-frz} and \textsc{CAP-ft} improve success rates by $8.2$ and $9.0$ percentage points, respectively, on RoboMimic, by $19.0$ and $13.8$ points on LIBERO, and by $5.8$ and $13.7$ points on RoboTwin when averaged over the eight quantizers. The relative performance of \textsc{CAP-frz} and \textsc{CAP-ft} depends on the tokenizer type. Averaged across the three benchmarks and both shared and level-specific configurations, \textsc{CAP-frz} improves over the matched baselines by $19.6$ percentage points for \emph{parallel} tokenizers (VQ and FSQ) and $2.4$ points for \emph{residual} tokenizers (RVQ and NRVQ). In comparison, \textsc{CAP-ft} yields improvements of $18.0$ and $6.3$ points, respectively.  Overall, \textsc{CAP-ft} outperforms \textsc{CAP-frz} by $3.9$ percentage points for \emph{residual} tokenizers, whereas \textsc{CAP-frz} leads by $1.6$ points for \emph{parallel} tokenizers.


\begin{takeaway}
    \textbf{Takeaway.} 1) CAP consistently outperforms standard prediction heads across both single-arm and bimanual manipulation. \textsc{CAP-ft} is more effective for \emph{residual} tokenizers, whereas \textsc{CAP-frz} is less sensitive to tokenizer structure and therefore a better choice for \emph{parallel} tokenizers.
    2) VQ-L delivers the strongest and most consistent policy performance across benchmarks, despite the recent prevalence of RVQ- and FSQ-based action tokenizers.
\end{takeaway}



\FloatBarrier

\begin{figure*}[t]
    \centering
    \input{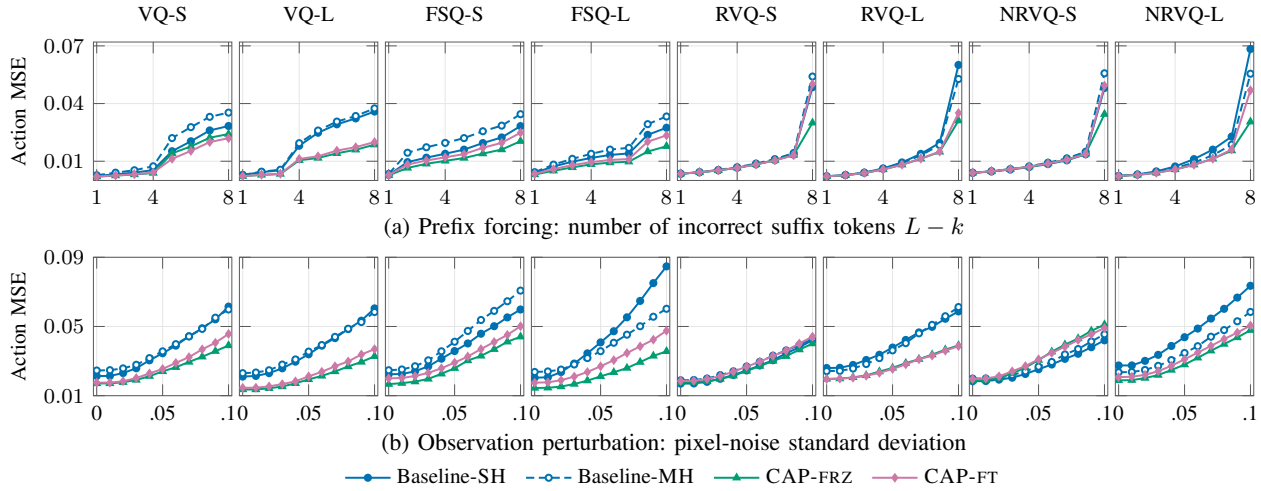}
    \pgfplotsset{
        action error split axis/.style={
            analysis axis,
            width=0.103\textwidth,
            height=0.103\textwidth,
            tick label style={font=\footnotesize},
            label style={font=\footnotesize},
            title style={font=\footnotesize,yshift=-2pt},
            scaled y ticks=false,
            yticklabel style={
                /pgf/number format/fixed,
                /pgf/number format/precision=3,
            },
        },
    }
    \begin{subfigure}{\linewidth}
        \begin{tikzpicture}
            \begin{groupplot}[
                group style={group size=8 by 1,horizontal sep=0.10cm,y descriptions at=edge left},
                  action error split axis,
                  xmin=1,xmax=8,xtick={1,4,8},
                  ymin=0,ymax=0.072,ytick={0.01,0.04,0.07},
                  legend style={analysis legend,font=\footnotesize},
                ]
                \nextgroupplot[title={VQ-S},ylabel={Action MSE}]
                \splitlegend{actionerrorlegend}
                \addplot[analysis Base,analysis S,forget plot] table[x=x,y=baseline_sh_s]{figures/analysis/data/prefix_vq.dat};
                \addplot[analysis Base,analysis MH,forget plot] table[x=x,y=baseline_mh_s]{figures/analysis/data/prefix_vq.dat};
                \addplot[analysis FRZ,analysis S,forget plot] table[x=x,y=cbfrz_s]{figures/analysis/data/prefix_vq.dat};
                \addplot[analysis FT,analysis S,forget plot] table[x=x,y=cbft_s]{figures/analysis/data/prefix_vq.dat};
                \nextgroupplot[title={VQ-L}]
                \addplot[analysis Base,analysis S,forget plot] table[x=x,y=baseline_sh_l]{figures/analysis/data/prefix_vq.dat};
                \addplot[analysis Base,analysis MH,forget plot] table[x=x,y=baseline_mh_l]{figures/analysis/data/prefix_vq.dat};
                \addplot[analysis FRZ,analysis S,forget plot] table[x=x,y=cbfrz_l]{figures/analysis/data/prefix_vq.dat};
                \addplot[analysis FT,analysis S,forget plot] table[x=x,y=cbft_l]{figures/analysis/data/prefix_vq.dat};
                \nextgroupplot[title={FSQ-S}]
                \addplot[analysis Base,analysis S,forget plot] table[x=x,y=baseline_sh_s]{figures/analysis/data/prefix_fsq.dat};
                \addplot[analysis Base,analysis MH,forget plot] table[x=x,y=baseline_mh_s]{figures/analysis/data/prefix_fsq.dat};
                \addplot[analysis FRZ,analysis S,forget plot] table[x=x,y=cbfrz_s]{figures/analysis/data/prefix_fsq.dat};
                \addplot[analysis FT,analysis S,forget plot] table[x=x,y=cbft_s]{figures/analysis/data/prefix_fsq.dat};
                \nextgroupplot[title={FSQ-L}]
                \addplot[analysis Base,analysis S,forget plot] table[x=x,y=baseline_sh_l]{figures/analysis/data/prefix_fsq.dat};
                \addplot[analysis Base,analysis MH,forget plot] table[x=x,y=baseline_mh_l]{figures/analysis/data/prefix_fsq.dat};
                \addplot[analysis FRZ,analysis S,forget plot] table[x=x,y=cbfrz_l]{figures/analysis/data/prefix_fsq.dat};
                \addplot[analysis FT,analysis S,forget plot] table[x=x,y=cbft_l]{figures/analysis/data/prefix_fsq.dat};
                \nextgroupplot[title={RVQ-S}]
                \addplot[analysis Base,analysis S,forget plot] table[x=x,y=baseline_sh_s]{figures/analysis/data/prefix_rvq.dat};
                \addplot[analysis Base,analysis MH,forget plot] table[x=x,y=baseline_mh_s]{figures/analysis/data/prefix_rvq.dat};
                \addplot[analysis FRZ,analysis S,forget plot] table[x=x,y=cbfrz_s]{figures/analysis/data/prefix_rvq.dat};
                \addplot[analysis FT,analysis S,forget plot] table[x=x,y=cbft_s]{figures/analysis/data/prefix_rvq.dat};
                \nextgroupplot[title={RVQ-L}]
                \addplot[analysis Base,analysis S,forget plot] table[x=x,y=baseline_sh_l]{figures/analysis/data/prefix_rvq.dat};
                \addplot[analysis Base,analysis MH,forget plot] table[x=x,y=baseline_mh_l]{figures/analysis/data/prefix_rvq.dat};
                \addplot[analysis FRZ,analysis S,forget plot] table[x=x,y=cbfrz_l]{figures/analysis/data/prefix_rvq.dat};
                \addplot[analysis FT,analysis S,forget plot] table[x=x,y=cbft_l]{figures/analysis/data/prefix_rvq.dat};
                \nextgroupplot[title={NRVQ-S}]
                \addplot[analysis Base,analysis S,forget plot] table[x=x,y=baseline_sh_s]{figures/analysis/data/prefix_nrvq.dat};
                \addplot[analysis Base,analysis MH,forget plot] table[x=x,y=baseline_mh_s]{figures/analysis/data/prefix_nrvq.dat};
                \addplot[analysis FRZ,analysis S,forget plot] table[x=x,y=cbfrz_s]{figures/analysis/data/prefix_nrvq.dat};
                \addplot[analysis FT,analysis S,forget plot] table[x=x,y=cbft_s]{figures/analysis/data/prefix_nrvq.dat};
                \nextgroupplot[title={NRVQ-L}]
                \addplot[analysis Base,analysis S,forget plot] table[x=x,y=baseline_sh_l]{figures/analysis/data/prefix_nrvq.dat};
                \addplot[analysis Base,analysis MH,forget plot] table[x=x,y=baseline_mh_l]{figures/analysis/data/prefix_nrvq.dat};
                \addplot[analysis FRZ,analysis S,forget plot] table[x=x,y=cbfrz_l]{figures/analysis/data/prefix_nrvq.dat};
                \addplot[analysis FT,analysis S,forget plot] table[x=x,y=cbft_l]{figures/analysis/data/prefix_nrvq.dat};
            \end{groupplot}
        \end{tikzpicture}
        \vspace{-5px}
        \caption{Prefix forcing: number of incorrect suffix tokens $L-k$}\label{fig:analysis-prefix-forcing}
    \end{subfigure}\\
    \begin{subfigure}{\linewidth}
        \begin{tikzpicture}
            \begin{groupplot}[
                group style={group size=8 by 1,horizontal sep=0.10cm,y descriptions at=edge left},
                action error split axis,
                xmin=0,xmax=0.1,xtick={0,0.05,0.1},xticklabels={0,.05,.1},
                ymin=0.01,ymax=0.09,ytick={0.01,0.05,0.09},
                ]
                \nextgroupplot[ylabel={Action MSE}]
                \addplot[analysis Base,analysis S,forget plot] table[x=x,y=baseline_sh_s]{figures/analysis/data/noise_vq.dat};
                \addplot[analysis Base,analysis MH,forget plot] table[x=x,y=baseline_mh_s]{figures/analysis/data/noise_vq.dat};
                \addplot[analysis FRZ,analysis S,forget plot] table[x=x,y=cbfrz_s]{figures/analysis/data/noise_vq.dat};
                \addplot[analysis FT,analysis S,forget plot] table[x=x,y=cbft_s]{figures/analysis/data/noise_vq.dat};
                \nextgroupplot
                \addplot[analysis Base,analysis S,forget plot] table[x=x,y=baseline_sh_l]{figures/analysis/data/noise_vq.dat};
                \addplot[analysis Base,analysis MH,forget plot] table[x=x,y=baseline_mh_l]{figures/analysis/data/noise_vq.dat};
                \addplot[analysis FRZ,analysis S,forget plot] table[x=x,y=cbfrz_l]{figures/analysis/data/noise_vq.dat};
                \addplot[analysis FT,analysis S,forget plot] table[x=x,y=cbft_l]{figures/analysis/data/noise_vq.dat};
                \nextgroupplot
                \addplot[analysis Base,analysis S,forget plot] table[x=x,y=baseline_sh_s]{figures/analysis/data/noise_fsq.dat};
                \addplot[analysis Base,analysis MH,forget plot] table[x=x,y=baseline_mh_s]{figures/analysis/data/noise_fsq.dat};
                \addplot[analysis FRZ,analysis S,forget plot] table[x=x,y=cbfrz_s]{figures/analysis/data/noise_fsq.dat};
                \addplot[analysis FT,analysis S,forget plot] table[x=x,y=cbft_s]{figures/analysis/data/noise_fsq.dat};
                \nextgroupplot
                \addplot[analysis Base,analysis S,forget plot] table[x=x,y=baseline_sh_l]{figures/analysis/data/noise_fsq.dat};
                \addplot[analysis Base,analysis MH,forget plot] table[x=x,y=baseline_mh_l]{figures/analysis/data/noise_fsq.dat};
                \addplot[analysis FRZ,analysis S,forget plot] table[x=x,y=cbfrz_l]{figures/analysis/data/noise_fsq.dat};
                \addplot[analysis FT,analysis S,forget plot] table[x=x,y=cbft_l]{figures/analysis/data/noise_fsq.dat};
                \nextgroupplot
                \addplot[analysis Base,analysis S,forget plot] table[x=x,y=baseline_sh_s]{figures/analysis/data/noise_rvq.dat};
                \addplot[analysis Base,analysis MH,forget plot] table[x=x,y=baseline_mh_s]{figures/analysis/data/noise_rvq.dat};
                \addplot[analysis FRZ,analysis S,forget plot] table[x=x,y=cbfrz_s]{figures/analysis/data/noise_rvq.dat};
                \addplot[analysis FT,analysis S,forget plot] table[x=x,y=cbft_s]{figures/analysis/data/noise_rvq.dat};
                \nextgroupplot
                \addplot[analysis Base,analysis S,forget plot] table[x=x,y=baseline_sh_l]{figures/analysis/data/noise_rvq.dat};
                \addplot[analysis Base,analysis MH,forget plot] table[x=x,y=baseline_mh_l]{figures/analysis/data/noise_rvq.dat};
                \addplot[analysis FRZ,analysis S,forget plot] table[x=x,y=cbfrz_l]{figures/analysis/data/noise_rvq.dat};
                \addplot[analysis FT,analysis S,forget plot] table[x=x,y=cbft_l]{figures/analysis/data/noise_rvq.dat};
                \nextgroupplot
                \addplot[analysis Base,analysis S,forget plot] table[x=x,y=baseline_sh_s]{figures/analysis/data/noise_nrvq.dat};
                \addplot[analysis Base,analysis MH,forget plot] table[x=x,y=baseline_mh_s]{figures/analysis/data/noise_nrvq.dat};
                \addplot[analysis FRZ,analysis S,forget plot] table[x=x,y=cbfrz_s]{figures/analysis/data/noise_nrvq.dat};
                \addplot[analysis FT,analysis S,forget plot] table[x=x,y=cbft_s]{figures/analysis/data/noise_nrvq.dat};
                \nextgroupplot
                \addplot[analysis Base,analysis S,forget plot] table[x=x,y=baseline_sh_l]{figures/analysis/data/noise_nrvq.dat};
                \addplot[analysis Base,analysis MH,forget plot] table[x=x,y=baseline_mh_l]{figures/analysis/data/noise_nrvq.dat};
                \addplot[analysis FRZ,analysis S,forget plot] table[x=x,y=cbfrz_l]{figures/analysis/data/noise_nrvq.dat};
                \addplot[analysis FT,analysis S,forget plot] table[x=x,y=cbft_l]{figures/analysis/data/noise_nrvq.dat};
            \end{groupplot}
        \end{tikzpicture}
        \vspace{-5px}
        \caption{Observation perturbation: pixel-noise standard deviation}\label{fig:analysis-observation-perturbation}
    \end{subfigure}
    \vspace{-5px}
    \pgfplotslegendfromname{actionerrorlegend}
    \caption{Action error under controlled \subref{fig:analysis-prefix-forcing} suffix errors and \subref{fig:analysis-observation-perturbation} observation perturbations.
    Columns correspond to the shared (S) and level-specific (L) variants of VQ, RVQ, NRVQ, and FSQ.
    Both Baseline-SH and Baseline-MH are shown; solid filled circles denote SH, and dashed open circles denote MH.}\label{fig:analysis}
    \vspace{-5px}
\end{figure*}

\subsection{Error Analysis}
\label{sec:analysis}
We first characterize how CAP, baseline-SH, and baseline-MH change token prediction errors and the resulting errors in action space at inference time, without further training.


\begin{table}[t]
  \centering
  \caption{Teacher-forced token accuracy (\%) on RoboMimic. Gray denotes the highest validation accuracy, while bold marks the policy with the best rollout success.}
  \label{tab:teacher-forced-accuracy}
  \scriptsize
  \renewcommand{\arraystretch}{0.92}
  \setlength{\tabcolsep}{1.7pt}
    \begin{tabular*}{\columnwidth}{
    @{\extracolsep{\fill}}
    l cccccccc
    @{}
    }
      \toprule
      & VQ-S & VQ-L & RVQ-S & RVQ-L & NRVQ-S & NRVQ-L & FSQ-S & FSQ-L \\
      \midrule
      Baseline-\textsc{SH} & 35.1 & 36.4 & \cellcolor{black!10}39.0 & 22.7 & \cellcolor{black!10}41.3 & 21.5 & 32.2 & 35.7 \\
      Baseline-\textsc{MH} & 32.8 & 35.7 & 36.2 & 23.3 & 38.0 & 22.7 & 30.8 & 33.7 \\
      \textsc{CAP-frz} & \cellcolor{black!10}\textbf{36.5} & \cellcolor{black!10}40.7 & 35.9 & 23.3 & 38.3 & \textbf{23.6} & \cellcolor{black!10}32.5 & \cellcolor{black!10}\textbf{36.7} \\
      \textsc{CAP-ft}  & 35.8 & \textbf{40.0} & \textbf{37.0} & \cellcolor{black!10}\textbf{25.1} & \textbf{38.8} & \cellcolor{black!10}23.8 & \textbf{31.9} & \cellcolor{black!10}36.7 \\
      \bottomrule
    \end{tabular*}%
  \vspace{-6px}
\end{table}

\textbf{Token accuracy does not explain CAP's gains.}
We examine whether CAP's rollout improvements can be explained by higher token prediction accuracy.
We measure next-token prediction accuracy under teacher forcing on the RoboMimic policy-validation split, conditioning on the observation and ground-truth token prefix.
As shown in Table~\ref{tab:teacher-forced-accuracy}, CAP has similar token accuracy to the baselines, and the policy with the highest validation token accuracy is the one that achieves the highest rollout success for only three of eight quantizers in Table~\ref{tab:main}.
Thus, validation token accuracy does not explain CAP's increase in rollout success rates.

\textbf{CAP yields more benign errors.}
Since token accuracy does not explain CAP's strong performance, we next test whether CAP makes token prediction errors that are more benign once decoded into actions. For each validation example, we force the first $k$ ground-truth tokens and autoregressively sample the remaining $L-k$ tokens, using rejection sampling at each suffix position to ensure that the sampled token is different from the ground-truth token. 
We repeat this procedure independently $16$ times per example at temperature $1$, yielding $16$ complete token sequences,
each with an exact Hamming distance of $L-k$.
We then measure the error between the decoded and ground-truth actions using $\ell_{\mathrm{rec}}$, the per-element MSE in raw action space defined in Eqn.~\ref{eq:reconstruction-error}.
Figure~\ref{fig:analysis-prefix-forcing} reports the mean $\ell_{\mathrm{rec}}$ as a function of $L-k$, the number of suffix token errors.
For six of the eight quantizers, both CAP variants consistently yield lower action error than both baselines, while action error is similar for the remaining two quantizers (RVQ-S and NRVQ-S).
Thus, when CAP policies make token prediction errors, the resulting decoded actions are closer to the ground-truth actions.

\textbf{CAP improves perturbation robustness.}
We finally test whether this benign error translates to greater robustness under perturbed observations.
We add zero-mean Gaussian noise with standard deviation $\sigma$ to each image pixel intensity, 
using the same noise realization across predictors, and greedily decode the resulting action tokens. 
Figure~\ref{fig:analysis-observation-perturbation} reports the mean $\ell_{\mathrm{rec}}$ as a function of $\sigma$. Across VQ, FSQ, RVQ-L, and NRVQ-L, both CAP variants consistently result in lower action error than both baselines for different amounts of Gaussian noise,
while the advantage is less pronounced for RVQ-S and NRVQ-S. 
These results highlight a practical consequence of the benign prediction errors induced by CAP: the resulting policies are more robust to perturbations in the visual observation.

%
%
\begin{takeaway}
    \textbf{Takeaway.}
    CAP does not improve the token prediction accuracy;
    instead, it is more robust to token prediction errors and observation perturbations, yielding more benign errors in action space.
\end{takeaway}

\subsection{Mechanism Analysis}
\label{sec:ablations}
We next investigate what gives rise to these differences. Through controlled ablations,
we examine three candidate mechanisms:
prediction-head parameterization, action-aware codebook structure, and representation learning in the policy backbone. We perform these ablations on both the matched baselines and our methods.
\subsubsection{Prediction-Head Parameterization}
We ask whether the benefit of CAP can be explained by the parameterization of its prediction head, rather than by its use of tokenizer code vectors.
We construct two ablated predictors: a \emph{distance head} (\textsc{DH}) and a \emph{low-rank predictor} (\textsc{LP}) that separately test distance-based scoring and low-rank parameterization.

The distance head ablation
replaces the tokenizer code vectors with randomly initialized, trainable prototypes, while retaining the same distance-based scoring rule (Eqn.~\ref{eq:distance-logits}).
In this way, it isolates the effect of the distance-based parameterization from the semantic structure supplied by the pretrained tokenizer.

The low-rank predictor ablation instead retains a standard learned prediction head but constrains it to a low-dimensional subspace.
We include this control because the tokenizer codebooks are substantially lower-rank than the learned baseline heads: across the eight tokenizer configurations, the mean codebook rank ranges from $3.2$ to $12.4$, whereas the corresponding policy-output rank ranges from $79.7$ to $111.3$ (approximately $9$--$32\times$ larger).
To test whether CAP's gains can be explained by this low-rank constraint alone, the low-rank predictor ablation factorizes its weight matrix as
$ W=BA,\quad A\in\mathbb R^{d_c\times d_h},\quad B\in\mathbb R^{K\times d_c}$
with no bias, where $d_h$ is the policy-backbone dimension, $K$ is the number of output classes, and $d_c$ matches the code-vector dimension.
Thus, \textsc{LP} imposes a rank-$d_c$ bottleneck.

\begin{table}[!t]
  \centering
  \caption{RoboMimic success rates for distance head (DH) and low-rank predictor (LP) prediction-head ablations.}
  \label{tab:ablations}
  \scriptsize
  \renewcommand{\arraystretch}{0.92}
  \setlength{\tabcolsep}{1.5pt}
  \resizebox{\columnwidth}{!}{%
    \begin{tabular}{lc cccc rr}
      \toprule
      & & \multicolumn{4}{c}{Success rate (\%)}
        & \multicolumn{2}{c}{Mean $\Delta$ (pp)} \\
      \cmidrule(lr){3-6}\cmidrule(lr){7-8}
      Method & Quant. & VQ & FSQ & RVQ & NRVQ
        & Baseline & \textsc{CAP-frz} \\
      \midrule
      \multirow{2}{*}{\textsc{DH}}
        & S & \sr{73.2}{1.5} & \sr{74.0}{1.5} & \sr{80.9}{1.3} & \sr{84.3}{1.2}
        & $-0.4$ & $-7.6$ \\
        & L & \sr{71.2}{1.5} & \sr{71.1}{1.5} & \sr{57.2}{1.6} & \sr{60.2}{1.6}
        & $-12.0$ & $-21.3$ \\
      \addlinespace[1pt]
      \multirow{2}{*}{\textsc{LP}}
        & S & \sr{64.6}{1.6} & \sr{68.8}{1.5} & \sr{74.1}{1.5} & \sr{72.2}{1.5}
        & $\hphantom{0}-8.6$ & $-15.7$ \\
        & L & \sr{71.2}{1.5} & \sr{74.4}{1.4} & \sr{43.7}{1.7} & \sr{42.0}{1.6}
        & $-19.1$ & $-28.4$ \\
      \bottomrule
    \end{tabular}%
  }
  \vspace{-6px}
\end{table}

Table~\ref{tab:ablations}
shows that neither the distance head nor the low-rank predictor ablation matches the performance of \textsc{CAP-frz} on RoboMimic.
The degradation of the distance head ablation is consistent with the fact that distance-based scoring does not fundamentally change the prediction function.
Let $h \in \mathbb{R}^{d_h}$ be the backbone output, and let $P\in\mathbb R^{d_c\times d_h}$ and $b\in\mathbb R^{d_c}$ denote the distance head's projection matrix and bias, respectively.
For prototype $v_k$, its logit is
\begin{equation*}
    \begin{aligned}
        \eta_k
        &=-\tau\lVert Ph+b-v_k\rVert_2^2 \\
        &=-\tau\lVert Ph+b\rVert_2^2
        +2\tau v_k^{\mathsf T}Ph
        +2\tau v_k^{\mathsf T}b
        -\tau\lVert v_k\rVert_2^2.
    \end{aligned}
\end{equation*}
The first term is shared by all tokens and therefore has no effect on the softmax probabilities, while the remaining terms form an affine function of $h$.
Thus, the distance head is itself an affine classifier, with its weights and biases parameterized by the prototypes $v_k$.
This explains why distance-based scoring alone does not recover the benefit of \textsc{CAP-frz}.

Likewise, the low-rank predictor ablation shows that low-rank parameterization alone does not account for the improvement. A low-rank factorization only restricts the expressivity of the prediction head.
Together, these results suggest that CAP's improvement does not arise from the prediction-head parameterization alone.



\subsubsection{Action-Aware Codebook Structure}

We test whether CAP benefits from the action-aware structure encoded by the tokenizer code vectors.
Unlike one-hot supervision, CAP relates token labels through their corresponding code vectors, 
whose distances reflect structure learned from actions.


We construct two ablations to test this effect.
The \emph{permuted codebook} (\textsc{PC}) randomly permutes the correspondence between frozen codebook prototypes and token labels while leaving the tokenizer and token embeddings unchanged.
This preserves the prototype set, including its norms, rank, and spectrum, but breaks the code--label alignment.
The \emph{soft-target} (\textsc{ST}) control retains vanilla prediction heads but uses the soft label $\pi_{n,l,k}\propto\exp(-d_{n,l,k}/T_l)$, 
where $d_{n,l,k}$ is the decoded Euclidean action error from substituting token $k$ at level $l$ while fixing all other ground-truth tokens. 
For each level, we choose a temperature $T_l$ such that the dataset-average entropy of these soft targets matches the dataset-average predictive entropy of \textsc{CAP-frz} at that level.
Soft-target control helps isolate whether the learned tokenizer representations are truly helpful, or whether CAP's performance is simply due to soft prediction targets that take into account action-space error.

\begin{table}[!t]
  \centering
  \caption{RoboMimic success rates for the permuted-codebook and soft-target controls.}
  \label{tab:token-space-geometry}
  \scriptsize
  \renewcommand{\arraystretch}{0.92}
  \setlength{\tabcolsep}{1.5pt}
  \resizebox{\columnwidth}{!}{%
    \begin{tabular}{lc cccc cc}
      \toprule
      & & \multicolumn{4}{c}{Success rate (\%)}
        & \multicolumn{2}{c}{Mean $\Delta$ (pp)} \\
      \cmidrule(lr){3-6}\cmidrule(lr){7-8}
      Method & Quant. & VQ & FSQ & RVQ & NRVQ
        & Baseline & \textsc{CAP-frz} \\
      \midrule
      \multirow{2}{*}{\textsc{PC}}
        & S & \sr{42.0}{1.7} & \hphantom{0}\sr{7.4}{0.9} & \sr{18.0}{1.3} & \sr{18.8}{1.3}
        & $-57.0$ & $-64.1$ \\
        & L & \sr{50.1}{1.6} & \hphantom{0}\sr{1.6}{0.4} & \hphantom{0}\sr{4.1}{0.7} & \hphantom{0}\sr{9.3}{1.0}
        & $-60.6$ & $-69.9$ \\
      \addlinespace[1pt]
      \multirow{2}{*}{\textsc{ST}}
        & S & \sr{79.8}{1.3} & \sr{76.7}{1.4} & \sr{86.7}{1.1} & \sr{84.6}{1.2}
        & $+3.4$ & $-3.7$ \\
        & L & \sr{78.6}{1.4} & \sr{77.6}{1.4} & \sr{71.3}{1.5} & \sr{74.2}{1.5}
        & $-1.5$ & $-10.8$ \\
      \bottomrule
    \end{tabular}%
  }
  \vspace{-6px}
\end{table}

As shown in Table~\ref{tab:token-space-geometry}, the permuted-codebook control
performs substantially worse than both \textsc{CAP-frz} and the baseline, 
showing that preserving the codebook statistics alone is insufficient when the code--label correspondence is broken.
In contrast, the soft-target control improves over one-hot training, indicating that information about the relative quality of alternative tokens is useful, 
although it still performs worse than \textsc{CAP-frz}. The effectiveness of the soft target may also depend on how its temperature is calibrated.
Together, these results suggest that the action-aware alignment between code vectors and token labels is an important contributor to CAP's success.

\subsubsection{Representation Learning}

We further ask whether CAP improves the representations learned by the policy backbone.
To isolate this effect, we freeze backbones trained with either the matched baseline or \textsc{CAP-frz}, and retrain only a new prediction head with the alternative parameterization. Specifically, the baseline-trained backbone is paired with a CAP head, while the CAP-trained backbone is paired with an unconstrained classifier. Both cases preserve the matched single- or multi-head structure.

\begin{table}[!t]
  \centering
  \caption{RoboMimic success rates for representation-learning ablations.}
  \label{tab:representation-transfer}
  \scriptsize
  \renewcommand{\arraystretch}{0.92}
  \setlength{\tabcolsep}{1.5pt}
  \resizebox{\columnwidth}{!}{%
    \begin{tabular}{lc cccc cc}
      \toprule
      & & \multicolumn{4}{c}{Success rate (\%)}
        & \multicolumn{2}{c}{Mean $\Delta$ (pp)} \\
      \cmidrule(lr){3-6}\cmidrule(lr){7-8}
      Backbone & Quant. & VQ & FSQ & RVQ & NRVQ
        & Baseline & \textsc{CAP-frz} \\
      \midrule
      \multirow{2}{*}{\textsc{CAP-frz}}
        & S & \sr{87.8}{1.1} & \sr{87.3}{1.1} & \sr{85.8}{1.2} & \sr{75.9}{1.4}
        & $+5.7$ & $-1.5$ \\
        & L & \sr{83.4}{1.3} & \sr{86.0}{1.2} & \sr{79.9}{1.3} & \sr{82.8}{1.3}
        & $+6.2$ & $-3.2$ \\
      \addlinespace[1pt]
      \multirow{2}{*}{Baseline}
        & S & \sr{79.2}{1.4} & \sr{77.3}{1.4} & \sr{81.2}{1.3} & \sr{66.3}{1.6}
        & $-2.5$ & $-9.7$ \\
        & L & \sr{84.2}{1.2} & \sr{75.6}{1.4} & \sr{66.9}{1.6} & \sr{59.0}{1.6}
        & $-5.5$ & $-14.8$ \\
      \bottomrule
    \end{tabular}%
  }
  \vspace{-6px}
\end{table}

Table~\ref{tab:representation-transfer}
reports the resulting RoboMimic success rates.
A frozen \textsc{CAP-frz} backbone paired with a newly trained vanilla prediction head consistently outperforms the corresponding baseline and recovers a substantial portion of the gap to the original \textsc{CAP-frz} policy.
In contrast, replacing the head of a frozen baseline backbone with a CAP head generally degrades performance.

This asymmetry suggests that the CAP head provides a useful training signal for shaping the upstream policy representation, rather than acting only as a better parameterization of the final classifier.

%
%
\begin{takeaway}
    \textbf{Takeaway.}
    CAP's gains are not explained by head parameterization alone, but are driven by action-aware codebook structure and improved policy backbone representations.  
\end{takeaway}

\subsection{Real-World Evaluation}
\label{sec:real-world}

We first trained four tokenizers on the combined Cup Hang and Cube Stack dataset. For each tokenizer, we trained two policies: a baseline with the single vanilla prediction head and our proposed \textsc{CAP-frz} policy.

%
%
\begin{table}[t]
  \centering
  \caption{Real-robot results: validation raw-action reconstruction MSE ($\downarrow$) and success rate (\%, $\pm$ bootstrap standard error) over 30 trials per task. For each tokenizer configuration, the best result is shown in bold.}
  \label{tab:real-world}
  \scriptsize
  \renewcommand{\arraystretch}{0.92}
  \setlength{\tabcolsep}{1.5pt}
    \begin{tabular*}{\columnwidth}{
        @{\extracolsep{\fill}}
        l c cc cc
        @{}
    }
      \toprule
      & & \multicolumn{2}{c}{Cup Hang} & \multicolumn{2}{c}{Cube Stack} \\
      \cmidrule(lr){3-4}\cmidrule(lr){5-6}
      Quantizer & Recon. ($10^{-4}$)$\downarrow$
        & \textsc{SH} & \textsc{CAP-frz}
        & \textsc{SH} & \textsc{CAP-frz} \\
      \midrule
      RVQ-S & 2.77
        & \sr{33.3}{8.6} & \bestsr{56.7}{9.0}
        & \sr{26.7}{8.1} & \bestsr{53.3}{9.1} \\
      RVQ-L & 2.04
        & \sr{10.0}{5.5} & \bestsr{36.7}{8.8}
        & \hphantom{0}\sr{3.3}{3.3} & \bestsr{53.3}{9.1} \\
      \addlinespace[1pt]
      FSQ-S & 2.18
        & \sr{33.3}{8.6} & \bestsr{66.7}{8.6}
        & \sr{46.7}{9.1} & \bestsr{56.7}{9.0} \\
      FSQ-L & 2.28
        & \sr{10.0}{5.5} & \bestsr{63.3}{8.8}
        & \sr{13.3}{6.2} & \bestsr{66.7}{8.6} \\
      \bottomrule
    \end{tabular*}%
  \vspace{-0.25em}
\end{table}
As shown in Table~\ref{tab:real-world}, \textsc{CAP-frz} consistently improves success rates across both tasks and all four tokenizers. The gains are particularly pronounced for the level-specific tokenizers. Averaged across the two tasks, the success rate of RVQ-L increases from $6.7 \% $ to $45.0 \%$, while that of FSQ-L increases from $11.7\%$ to $65.0\%$.

Reconstruction error alone does not predict downstream policy performance. Although RVQ-L achieves the lowest reconstruction error among the four tokenizers, it yields one of the weakest baseline policies.
This suggests that reconstruction quality is not sufficient for assessing the suitability of an action tokenizer for policy learning.

\section{Conclusion}
\label{sec:conclusion}

We study an overlooked interface in tokenized robot policies: after an action tokenizer is trained, its discrete indices are retained as prediction targets while the corresponding code representations are typically discarded. 
We introduce \emph{Codebook-Aligned Prediction} (CAP), which reuses these representations as class prototypes and aligns the policy prediction heads with the tokenizer's quantization scheme. 
Across four quantizer families, three simulation benchmarks, and two real-world manipulation tasks, CAP consistently improves task performance over standard classification heads.
Our analyses provide insights into the reasons for this improvement---
CAP does not simply improve token prediction accuracy; 
instead, its prediction errors are more benign after decoding into action space and its policies are more robust to observation perturbations.  Controlled ablations show that these gains cannot be explained by distance-based scoring or low-rank head parameterization alone. 
Instead, preserving the action-aware alignment between code vectors and token labels is important, possibly because this improves the representations learned by the policy backbone.

Our results also motivate broader questions about the design of action tokenizers. Future work should revisit which quantization schemes are most appropriate for downstream tasks. 
We also plan to evaluate these tokenizer design choices and CAP across a broader range of tasks and models, including VLA policies and world models.

\balance
\bibliographystyle{IEEEtranN}
\bibliography{references}
\end{document}